%% file: main.tex
\documentclass[10pt,twocolumn,letterpaper]{article}

\usepackage[pagenumbers]{cvpr} % To force page numbers, e.g. for an arXiv version

\usepackage[utf8]{inputenc} % allow utf-8 input
\usepackage{graphicx}
\usepackage{amsmath}
\usepackage{amssymb}
\usepackage{booktabs}
\usepackage{times}
\usepackage{epsfig} 
\usepackage{makecell}
\usepackage{times}
\usepackage{silence}
\usepackage{epstopdf}
\usepackage{color, colortbl}
\usepackage{float}  
\usepackage{caption}
\usepackage{multirow} 
\usepackage{xcolor}
\usepackage{url}
\usepackage{cite}
\usepackage{placeins}
\usepackage[group-separator={,}]{siunitx}
\usepackage[utf8]{inputenc}
\usepackage[pagebackref,breaklinks,colorlinks]{hyperref}

\newcommand{\vspaceunderfig}{\vspace{-0.2cm}}
\newcommand{\vspaceundertab}{\vspace{-.2cm}}
\newcommand{\vspaceunderfigext}{\vspace{-0.7cm}}

\newcommand{\comment}[1]{}

\definecolor{LightCyan}{rgb}{0.88,1,1}
\definecolor{Gray}{gray}{0.9}
\definecolor{Pink}{rgb}{1, 0, 1}

\usepackage[pagebackref,breaklinks,colorlinks]{hyperref}

\usepackage[capitalize]{cleveref}
\crefname{section}{Sec.}{Secs.}
\Crefname{section}{Section}{Sections}
\Crefname{table}{Table}{Tables}
\crefname{table}{Tab.}{Tabs.}

\def\cvprPaperID{4268} % *** Enter the CVPR Paper ID here
\def\confName{CVPR}
\def\confYear{2022}

\begin{document}

% \title{SeMask: Semantically Masking Transformer Backbones for Effective Semantic Segmentation}
\title{SeMask: Semantically Masked Transformers for Semantic Segmentation}

%\title{SeMask: Semantically Masking Backbones for Effective Semantic Segmentation}

\author{Jitesh Jain\textsuperscript{1,2,3}\thanks{Work done during an internship at Picsart AI Research (PAIR).} \quad Anukriti Singh\textsuperscript{2} \quad Nikita Orlov\textsuperscript{1} \\ \quad Zilong Huang\textsuperscript{2} \quad Jiachen Li\textsuperscript{1,2} \quad Steven Walton\textsuperscript{1,2} \quad Humphrey Shi\textsuperscript{1,2}\\
{\small \textsuperscript{1}Picsart AI Research (PAIR) \qquad \textsuperscript{2}SHI Lab @ University of Oregon \& UIUC \qquad \textsuperscript{3}IIT Roorkee}}

\maketitle

%\blfootnote{Work done during an internship at Picsart AI Research (PAIR).}
\input{parts/00_abstract}
\input{parts/01_intro}
\input{parts/02_rel_work}
\input{parts/03_method}
\input{parts/04_experiments}
\input{parts/05_conclusion}

%%%%%%%%% REFERENCES
\bibliographystyle{ieee_fullname}
\bibliography{main}

\input{supplementary/supplementary}

\end{document}

%% file: parts/00_abstract.tex
\begin{abstract}

Finetuning a pretrained backbone in the encoder part of an image transformer network has been the traditional approach for the semantic segmentation task. However, such an approach leaves out the semantic context that an image provides during the encoding stage. This paper argues that incorporating semantic information of the image into pretrained hierarchical transformer-based backbones while finetuning improves the performance considerably. To achieve this, we propose SeMask, a simple and effective framework that incorporates semantic information into the encoder with the help of a semantic attention operation. In addition, we use a lightweight semantic decoder during training to provide supervision to the intermediate semantic prior maps at every stage. Our experiments demonstrate that incorporating semantic priors enhances the performance of the established hierarchical encoders with a slight increase in the number of FLOPs. We provide empirical proof by integrating SeMask into Swin Transformer and Mix Transformer backbones as our encoder paired with different decoders. Our framework achieves a new state-of-the-art of $58.25\%$ mIoU on the ADE20K dataset and improvements of over $3 \%$ in the mIoU metric on the Cityscapes dataset.  The code and checkpoints are publicly available at~
\href{https://github.com/Picsart-AI-Research/SeMask-Segmentation}{https://github.com/Picsart-AI-Research/SeMask-Segmentation}.
\end{abstract}

%% file: parts/01_intro.tex
\section{Introduction}

Semantic Segmentation aims to perform dense prediction for labeling each pixel in an image corresponding to the class that the pixel represents. Transformer-based vision networks \cite{vit,deit} have outperformed Convolutional Neural Networks on the image-classification task \cite{imagenet}. In modern times, transformer backbones have shown impressive performance when transferred to downstream tasks like semantic segmentation \cite{beit, fapn, swin-T}. 

Most of the architectural designs in vision transformers approach the problem in either of the two ways: (i) Use an existing pretrained backbone as an encoder and transfer it to downstream tasks using pre-existing standard decoders such as, Semantic FPN \cite{sem-fpn} or UperNet \cite{upernet}; OR (ii) design a new encoder-decoder network where the encoder is pretrained on ImageNet for the semantic segmentation task. Both of these ways, as mentioned earlier, involve finetuning the encoder backbone on the segmentation task. Finetuning from a large-scale dataset help early attention layers to incorporate local information at lower layers of the transformers \cite{vit-cnn}. However, it can still not harness the semantic context during finetuning due to the relatively smaller size of the dataset and a change in the number and nature of semantic classes from classification to the segmentation task. Hierarchical vision transformers \cite{segformer,swin-T} tackle the problem with progressive downsampling of features along the stages, although they still lack the semantic context of the image.

%The introduction of hierarchical structure in encoders of vision transformers try to handle this by using small-sized image patches in the start of the network and gradually merging neighboring patches in deeper layers. Even though, hierarchical transformers are most suited for dense prediction task like segmentation it still lacks the semantic context of the image. 

\begin{figure}
  \centering
\includegraphics[width=\linewidth]{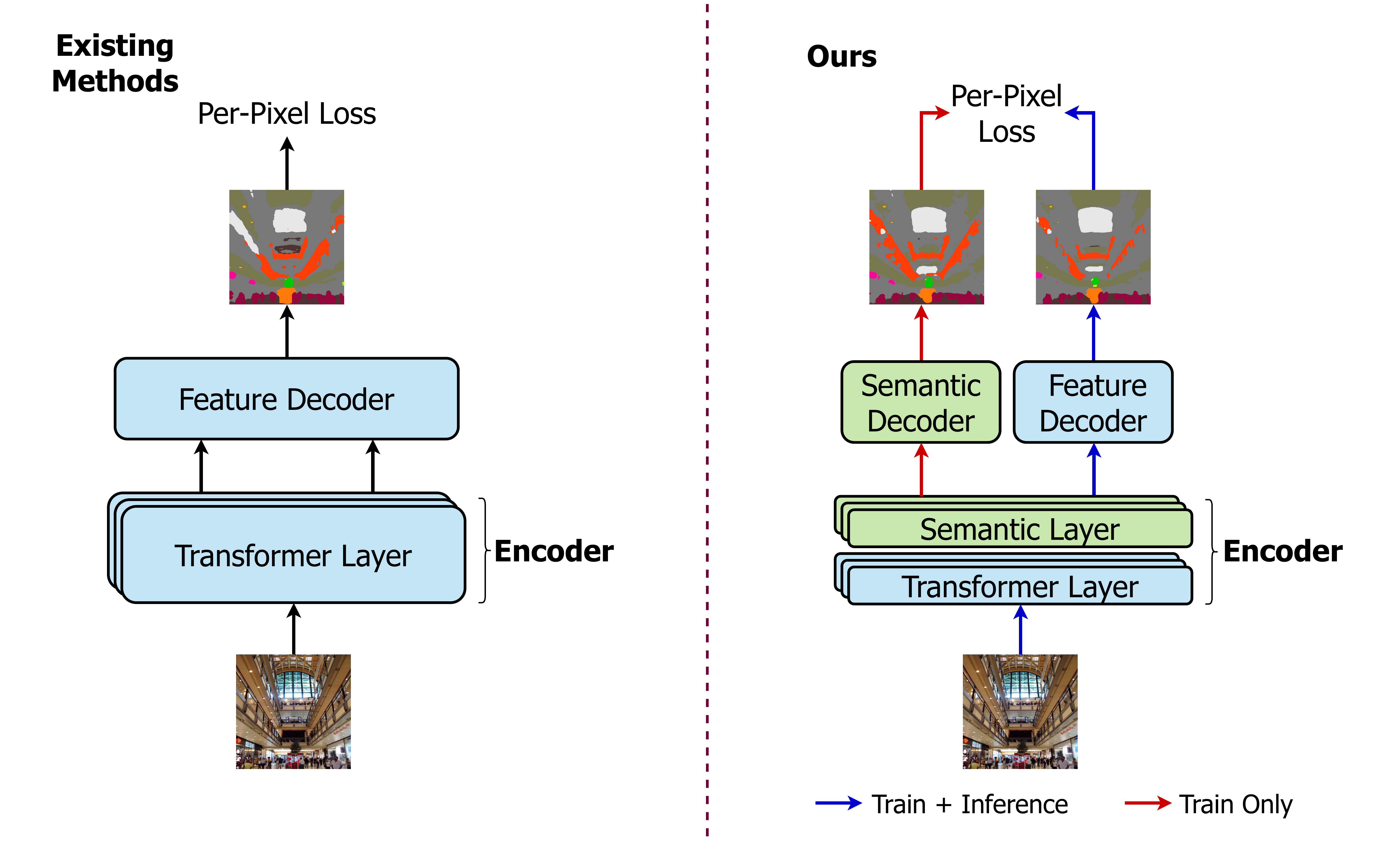}
  \vspaceunderfigext
  \caption{
      Comparison between popular transformer-based network for segmentation (left) and SeMask (right). In contrast to most existing methods (\cite{swin-T} in above figure) that directly use the pretrained backbones without any changes, SeMask uses semantic priors in the encoder backbones by adding an additional semantic layer; this simple change significantly improves performance.
      }
  \label{fig:intro}
\end{figure}

Liu et al. \cite{swin-T} introduced the Swin Transformer, which constructs hierarchical feature maps making it compatible as a general-purpose backbone for major downstream vision tasks. \cite{twins} proposed to use two attention: globally sub-sampled and locally sub-samples on top of PVT \cite{pvt} and CPVT \cite{cpvt} for effective segmentation. Xie et al. \cite{segformer} further modified the hierarchical transformer encoder by making it free from positional-encoding and thus robust to different resolutions as generally found in the segmentation task. All these works modified the encoders to make them work better for downstream tasks like segmentation and achieved success to an impressive extent. Still, they did not pay attention to capturing the semantic-level contextual information of the whole image. A lack of semantic contextual information leads to sub-optimal segmentation performance, especially in the case of small objects where those get merged with the boundaries of the larger categories, leading to wrong predictions. Recently, \cite{segmenter} tried to tackle this issue by designing a pure transformer-based decoder that jointly processes the patch and class embedding. However, it does not perform efficiently for tiny variants and fails with hierarchical architectures leading to sub-optimal performance when used with major transformer backbones like Swin \cite{swin-T}, and Twins \cite{twins} transformers.

%However, they make the model very specific by using ViT as their encoder and only modifying the decoder with class knowledge. The decoder is inflexible with hierarchical encoders (\todo{exp on base model}). 

Jin et al. in \cite{isnet} proposed ISNet to model the image level contextual information along with semantic level contextual information by introducing the SLCM and ILCM modules in the decoder structure. However there is still a caveat: ISNet is a CNN based method and only focuses on the decoder part of the network, leaving out the encoder unchanged.

% . We investigate whether providing a semantic-level information in the encoder helps to better group the pixels in their respected category

To address the issues mentioned above, we propose the SeMask framework that incorporates semantic information into hierarchical vision transformer architectures and augments the global feature information captured by the transformers with the semantic context. The existing frameworks formulate the architecture as an encoder-decoder structure with transformers pretrained on ImageNet \cite{imagenet} acting as the encoders and using a specialized decoder for semantic segmentation. In contrast to directly using the hierarchical transformers as a backbone, we insert a Semantic Layer after the Transformer Layer at each stage in the backbone, giving us the SeMask version of the backbone as illustrated in \cref{fig:intro}. We use a lightweight semantic decoder to accumulate the semantic maps from all the stages, and a standard decoder like Semantic-FPN \cite{sem-fpn} for the main per-pixel prediction. The added semantic modeling with feature modeling throughout the encoder helps us improve the performance of the semantic segmentation task. In \cref{sec:experiments}, we integrate the proposed SeMask block into the Swin Transformer \cite{swin-T} and Mix Transformer \cite{segformer} backbones. Our experimental results show considerable improvement in semantic segmentation for both backbones on two different datasets. To summarize, our contributions are three fold:  

%With SeMask we aim to introduce semantic information about the category of the pixels in the encoder part of the transformer. We use a pretrained  
% We use semantic mask to help attend class category as a whole instead of individual features. This way, the model learns class importance along with feature importance which helps to improve the performance of the encoder during training.\todo{mention tiny model improvement} \todo{talk about figure 1}
%In a hierarchical vision transformer backbone, SeMask demonstrates high performance for semantic segmentation. 

\begin{itemize}
    \item To the best of our knowledge, we are the first to study the effect of adding semantic context to pretrained transformer backbones for the semantic segmentation task. Furthermore, we introduce a SeMask Block which can be plugged into any existing hierarchical vision transformer. We provide empirical evidence by integrating SeMask into Swin-transformer~\cite{swin-T} and Mix-Transformer~\cite{segformer}, and achieving considerable performance improvement.
    %Along with the global MHSA, SeMask Block uses a \textit{semantic attention} operation to predict a semantic prior guiding the encoder's image-level feature modeling process.
    \item We also propose to use a simple semantic decoder for aggregating the semantic priors from different stages of the encoder. The semantic priors receive supervision from the ground truth using a per-pixel cross-entropy loss.
    \item Lastly, we provide an in-depth analysis of the SeMask Block's effect on two different datasets: ADE20K and Cityscapes. We achieve the new state-of-the-art performance on the ADE20K dataset and an improvement above $3\%$ on the Cityscapes dataset.
\end{itemize}

%% file: parts/02_rel_work.tex
\section{Related Work}
\subsection{Semantic Segmentation}
Semantic segmentation broadly formulates to a dense per-pixel classification task. The seminal work of FCN \cite{fcn} introduced the use of deep CNNs, removing fully connected layers to tackle the segmentation task. Several following works \cite{segnet, unet, refinenet} were built upon the same idea of using the encoder-decoder architecture. \cite{deeplabv1} introduced the use of atrous convolutions inside the DCNN to tackle the signal downsampling issue. Later, various works focused on the aggregating long-range context in the final feature map: ASPP \cite{deeplabv2, deeplabv3, deeplabv3+} uses atrous convolutions with different dilation rates; PPM \cite{pspnet} uses pooling operators with different kernel sizes.

The recent DCNN based models focus on efficiently aggregating the hierarchical features from a pretrained backbone based encoder with specially designed modules: \cite{attanet, lednet, hrnet-ocr} introduce attention modules in the decoder; \cite{ccnet, dualnet} use different forms of non-local blocks \cite{nonlocal}; \cite{sfnet} proposes a novel FAM module to solve the misalignment issue using semantic flow; AlignSeg \cite{Alignseg} proposes aligned feature aggregation module and aligned context modeling module to make contextual features be better aligned. \cite{shelfnet} uses a segmentation shelf for better information flow. In this work, we also follow the established direction to use a pretrained backbone and aggregating the hierarchical features \cite{swin-T} using the Semantic-FPN \cite{sem-fpn} decoder.

\begin{figure*}[ht!]
    \centering
    \includegraphics[width=\linewidth,height=11cm]{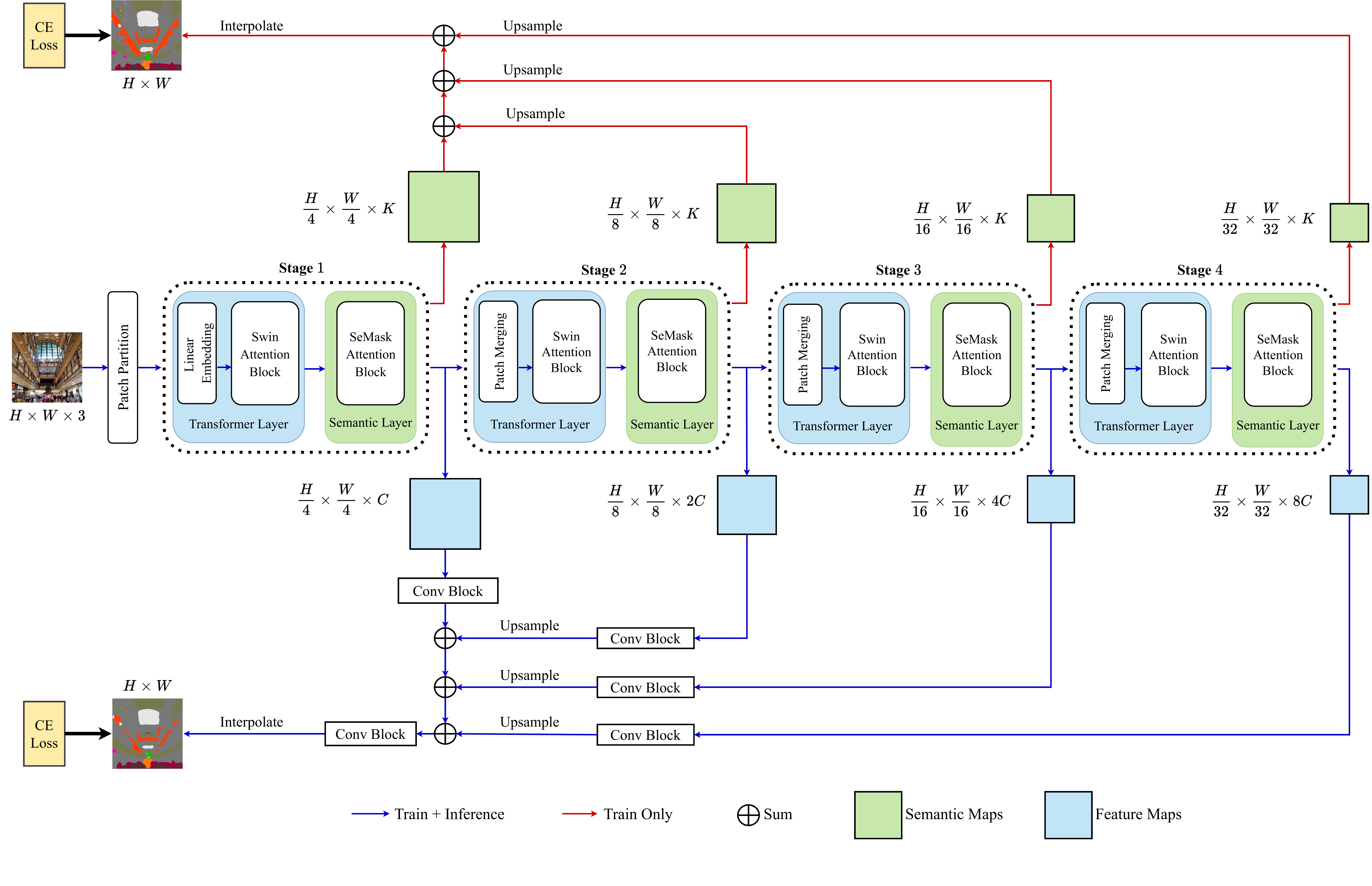}
    \vspaceunderfig
    \vspace{-.35cm}
    \caption{\textbf{SeMask Swin Semantic FPN Framework:} 
    We add a Semantic Layer with $N_S$ SeMask Blocks (\cref{fig:semask-block}) after the Swin Transformer Layer to capture the semantic context in the encoder network. The Semantic Maps from the Semantic Layers at each stage are aggregated using a simple Upsample + Sum operation and passed through a weighted CE Loss to supervise the semantic context.
        }
    \label{fig:semask}
\end{figure*}

\subsection{Transformers for Segmentation}
After being heavily used in Natural Language Processing field, transformer \cite{vaswani2017attention} based models have gained popularity for various computer vision tasks since the introduction of ViT \cite{vit} for image classification \cite{vit, perceiver, deit, cctv1}. SETR used ViT \cite{vit} as an encoder and two decoders based upon progressive upsampling and multi-level feature aggregation. SegFormer \cite{segformer} proposed to use a hierarchical pyramid vision transformer network as an encoder with an MLP based decoder to obtain the segmentation mask. 
Segmenter \cite{segmenter} designed mask transformer as a decoder, which uses learnable class-map tokens to enhance decoding performance. 
MaskFormer \cite{maskformer} defines the problem of per-pixel classification from a mask classification point of view, creating an all-in-one module for all segmentation tasks. Mask2Former \cite{mask2former} further evolves masked attention to solve panoptic, instance and semantic segmentation tasks in one framework. Most recent transformer-based segmentation frameworks \cite{swin-T, cswin-T} are based on finetuning a pretrained hierarchical backbone as an encoder, and standard decoders like Semantic-FPN and UperNet \cite{sem-fpn, upernet} to the segmentation task. In this work, we follow the same paradigm and, in addition, propose a framework to enhance the finetuning ability of the pretrained vision transformer backbone. Note that there is also recent concurrent work like SwinV2 \cite{swinv2} that reaches new state-of-the-art performance on ADE20k benchmark by using improved and giant backbones (e.g. SwinV2-G with 3.0 billion parameters). That is out of the scope of this work and we follow the current practice mainly based on Swin-L backbone. Theoretically, we can get even better performance if we apply our approach to such giant models.

\subsection{Semantic Context in Segmentation}

Zhang et al. proposed the Context Encoding Module in \cite{encnet} which captures the global semantic context along with a feedback loop to balance the importance of classes in the features extracted by a ResNet backbone \cite{resnet}. 
More recently, \cite{isnet, mining-context} focus on capturing and integrating the semantic-level contextual information along with the image-level context with specially designed decoders which shows significant improvement in DCNN based methods. 
Each of these works captures the semantic context after the encoding stage based on the extracted features and not the encoder's ability to capture the semantic features.

In this work, we argue that semantic information is lost during the encoding stage and hence, propose a framework to capture semantic information which can be plugged into any pretrained vision transformer backbone network.

%% file: parts/03_method.tex
\section{Method}
%\subsection{Overall Architecture}
An overview of our architecture with Swin-Transformer~\cite{swin-T} backbone is shown in \cref{fig:semask}. The RGB input image, size $H \times W \times 3$, is first split into non-overlapping patches of size $4 \times 4$. The smaller size of the patch supports dense prediction in segmentation. These patches act as tokens and are given as input to the hierarchical vision transformer encoder, which is the Swin-Transformer \cite{swin-T} in our architecture. The encoding step consists of four different stages of hierarchical feature modeling. Every stage during the encoding step consists of two layers: The transformer layer, which is $N_A$ number of Swin Transformer blocks (\cref{fig:swin-block}) stacked together and Semantic Layer with $N_S$  number of SeMask Attention blocks (\cref{fig:semask-block}). We collectively refer to the Transformer Layer and Semantic Layer at each stage as our SeMask Block. The patch tokens pass through each stage at $\{\frac{1}{4}, \frac{1}{8}, \frac{1}{16}, \frac{1}{32}\}$ of the original image resolution for the feature maps and intermediate semantic-prior maps extraction.

In the encoder part of the network, the Semantic Layer takes in features from the Transformer Layer as inputs and returns the intermediate semantic-prior maps and semantically masked features (\cref{fig:semask-block}). When we plug the SeMask Attention Block into other hierarchical vision transformers, the Transformer Layer consists of attention blocks corresponding to the specific backbone, like Efficient-Self Attention-based Transformer Layer for the Mix Transformer~\cite{segformer} backbone. The semantically masked features from each stage are aggregated using the semantic-FPN \cite{sem-fpn} decoder for producing the final dense-pixel prediction. Moreover, the semantic-prior maps from all the stages are aggregated using a lightweight upsample \& sum operation-based semantic decoder to predict the semantic-prior for the network during training. Both decoders' outputs are supervised using a weighted per-pixel cross-entropy loss. These additional semantic-prior maps greatly assist the feature extraction and eventually improve the performance on the semantic segmentation task.

\subsection{SeMask Encoder}
\begin{figure}[ht!]
\centering
\subfloat[Swin Attention Block.]{
	\label{fig:swin-block}
	\includegraphics[width=\linewidth]{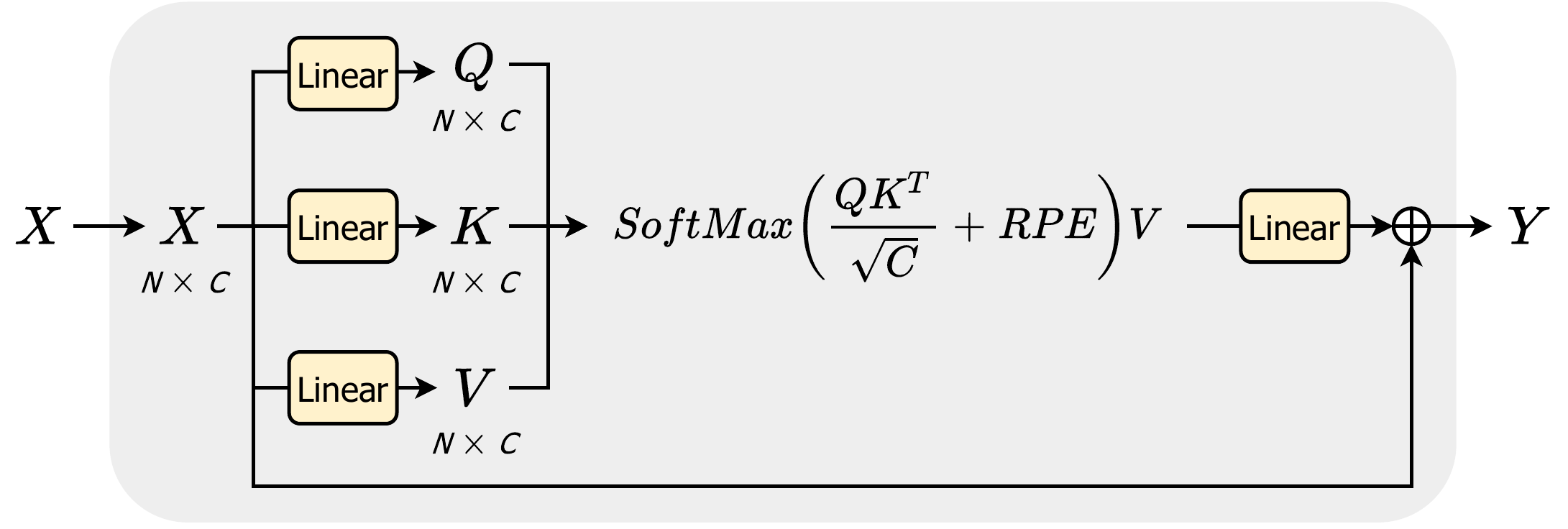} }
	
\subfloat[SeMask Attention Block.]{
	\label{fig:semask-block}
	\includegraphics[width=\linewidth]{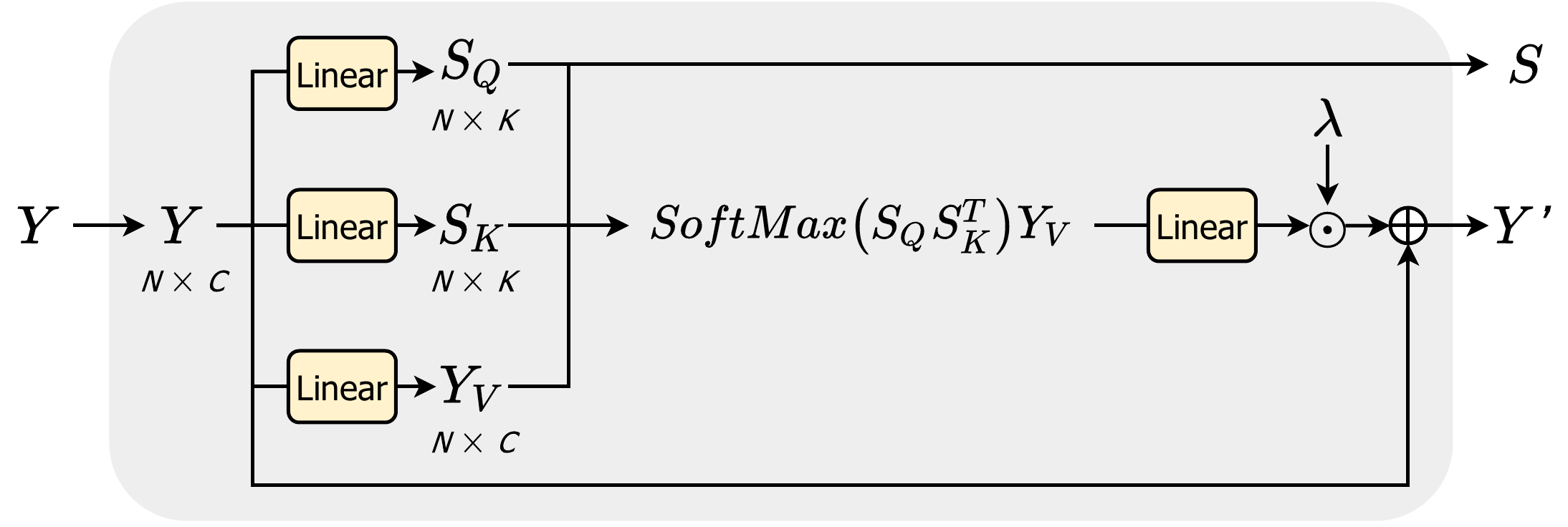} }

\caption{\textbf{Attention Blocks.} $N_A$ Shifted Window Self Attention Blocks, shown in \cref{fig:swin-block}, are stacked inside each Transformer Layer and $N_S$ SeMask Attention Blocks, shown in \cref{fig:semask-block}, are stacked inside each Semantic Layer at every stage (\cref{fig:semask}). The output, $Y$, from the last Swin Attention Block, is fed to the first SeMask block in the Semantic Layer.}
\label{fig:attn-block}
\end{figure}

Each stage in our encoder consists of two layers: the Transformer Layer and the Semantic Layer. The transformer layer is composed of $N_A$ Swin Transformer blocks stacked to extract image-level context information from the image. The semantic layer contains $N_S$ SeMask Attention blocks stacked together to decouple semantic information from the features, producing semantic-priors and then updating the features with guidance from these semantic-prior maps. \\
\textbf{Transformer layer.} For the transformer layer, we adapt the hierarchical structure of Swin Transformer \cite{swin-T} which constructs hierarchical feature maps and has linear computational complexity to the image resolution.
Before feeding the RGB image into the transformer layer in the first stage, we split it into non-overlapping patches of size is $4 \times 4 \times 3 = 48$. 
%The features of these patches are a concatenation of raw RGB pixel values. 
The first stage in the encoder has a linear embedding layer to change the feature dimension of the patch tokens. Inside each transformer layer, there are $N_A$ shifted window attention blocks (\cref{fig:swin-block}) that have linear computation complexity along with cross-window connections to handle non-overlapping regions, making the design effective for image-level feature modeling. For a hierarchical representation, we shrink our feature maps from $\frac{H}{4}\times \frac{W}{4}$ to $\frac{H}{8}\times \frac{W}{8}$ by patch merging layers for the next stage. This patch merging is iterated for the next stages to obtain a hierarchical feature map, with a  resolution of $\frac{H}{2^{i+1}}+\frac{W}{2^{i+1}}\times C_{i}$ where $i \in\left\{ 1, 2, 3, 4 \right\}$. $X$ represents the input features inside the transformer layer block. And for computing self-attention in the transformer layer, $X$ is transformed into: $Q, K, V$ which are \emph{query}, \emph{key} and \emph{value} matrices with same dimension of $N \times C$. Based on swin transformer, we also follow \cite{unilmv2, relation-net, local-relation-net, jmlr, swin-T} to include a relative position embedding (RPE) where $RPE \in \mathbb{R}^{N \times N}$ and $N = M \times M$ is the length of the sequence with $M=$ window size. The attention inside the Transformer Layer is calculated as:
\begin{equation}
\label{eq:att}
    \text{Attention}(Q, K, V) = \text{SoftMax}\left(\frac{QK^T}{\sqrt{C}}+RPE\right)V 
\end{equation}

The resulting feature $Y$ from the Transformer Layer after the last Swin Transformer block then acts as an input to the subsequent semantic layer in the same stage as shown in \cref{fig:attn-block}.

\textbf{Semantic Layer.}  The Semantic Layer follows the Transformer Layer at each stage of our hierarchical vision transformer. Unlike the Transformer Layer, the Semantic Layer's significance is in modeling the semantic context, which is used as a prior for calculating a segmentation score to update the feature maps based on guidance from the semantic nature present in the image. 
% While the transformer layer operates on the whole image to capture the contextual information, the semantic layer operates to aggregate the contextual information within individual class embedding.
Inside each semantic layer, there are $N_S$ SeMask attention blocks (\cref{fig:semask-block}). Inspired by the shifted window-based division of the tokens for efficient computation cost, we also divide the input to our SeMask blocks into windows with cross-window connections before calculating the segmentation score using a single-head self-attention operation. The SeMask block is responsible for capturing the semantic context in our encoder. It updates the features from the transformer layer from the segmentation score  providing guidance and giving a semantic-prior map for efficient supervision of the semantic modeling during training. SeMask attention block divides the features $Y$ from the preceding transformer layer into three entities: Semantic Query ($S_Q$), Semantic Key ($S_K$), and Feature Value ($Y_V$). We get $S_K$ and $S_Q$ by projecting the features onto the semantic space. The dimension of both $S_Q$ and $S_K$ is $N \times K$ where $K$ is equal to the number of classes, and the dimension of $Y_V$ is $N \times C$ where $C$ is the embedding dimension, $N = M \times M$ is the length of the sequence with $M=$ window size which we set as equal to that used inside the transformer layer. $S_Q$ returns the semantic map, and a segmentation score is calculated using $S_K$ and $S_Q$. The score is passed through a softmax and is used to update $Y_V$ as shown in \cref{fig:semask-block}. This SeMask attention equation is expressed as follows:
\begin{equation}\label{eq:sem-att}
    \text{Score}(S_Q, S_K, Y_V) = \text{SoftMax}(S_{Q}S^{T}_{K})Y_{V}
\end{equation}

We perform a matrix multiplication between the feature values and the segmentation score. The matrix product is later passed through a linear layer and multiplied with a learnable scalar constant $\lambda$, used for smooth finetuning. After a residual connection \cite{resnet}, we finally get the modified features, rich with semantic information which we call the \textbf{Se}mantically \textbf{Mask}ed features. The semantic queries $S_Q$ are later used to predict the semantic-prior map.

\subsection{Decoder}
\label{sec:decoder}
We use two decoders to aggregate the features and the semantic-prior maps respectively from the different stages in the encoder.

For aggregating the semantically masked features, we employ the popular Semantic-FPN decoder \cite{sem-fpn}. The Semantic-FPN fuses the features from different stages with a series of convolution, bilinear upsampling, and sum operations, making it efficient and straightforward as a segmentation decoder for our purpose.
In addition, we use a lightweight semantic decoder during training to provide ground truth supervision to the semantic-prior maps at every stage of the encoder. As the semantic-prior maps have the channel dimension of $K$ in each stage, we only employ a series of upsampling and sum operations to aggregate the maps with $K$ being equal to the number of classes in the dataset. Lastly, the output from both the decoders is up-scaled $\times 4$ to the resolution of the original image for the final predictions as shown in \cref{fig:semask}.

% The information from all levels of the FPN pyramid is merged into a single output. Starting from the deepest FPN level (at $1/32$ scale), we perform three upsampling stages to yield a feature map at $1/4$ scale. Each upsampling stage consists of 3×3 convolution, group norm \cite{group-norm}, ReLU, and $2 \times$ bilinear upsampling. This strategy is repeated for FPN scales $1/16$, $1/8$, and $1/4$ (with progressively fewer upsampling stages). The result is a set of feature maps at the same $1/4$ scale, which are then element-wise summed. A final $1 \times 1$ convolution, $4 \times$ bilinear upsampling, and softmax are used to generate the per-pixel class categories at the original image resolution.
% A second decoder is applied only at the time of training. It is used to provide supervision to intermediate semantic maps after each SeMask attention block. All the semantic maps are aggregated with upsampling from each stage in the transformer as shown in the upper region in \cref{fig:semask}. Finally, the output from both the decoders is upscaled to the resolution of the original image. \toedit{direct influence on the quality of segmentic map}

\subsection{Loss function}
To train our model's parameters, we calculate the total loss $\mathcal{L}_{T}$ as a summation of two per-pixel cross-entropy losses: $\mathcal{L}_{1}$ and $\mathcal{L}_{2}$. The loss $\mathcal{L}_{1}$ is calculated on the main prediction from the Semantic-FPN decoder and loss $\mathcal{L}_{2}$ is calculated on the semantic-prior prediction from our light-weight decoder. $\mathcal{F}$ contains the main prediction of the network and $\mathcal{S}$ denotes the semantic-prior prediction. We define our losses on $\mathcal{F}$ and $\mathcal{S}$ as follows: 
\begin{equation}
   \mathcal{L}_{1} = \frac{1}{H \times W} \sum_{i, j} \mathcal{L}_{ce} \left(\mathcal{F}_{[*, i, j]}, ~\S \left(\mathcal{GT}_{[ij]}\right)\right).
\end{equation}

\begin{equation}
   \mathcal{L}_{2} = \frac{1}{H \times W} \sum_{i, j} \mathcal{L}_{ce} \left(\mathcal{S}_{[*, i, j]}, ~\S \left(\mathcal{GT}_{[ij]}\right)\right).
\end{equation}

\begin{equation}\label{eq18}
   \mathcal{L_T} = \mathcal{L}_{1} + \alpha\mathcal{L}_{2}
\end{equation}

Here,  $\S$ denotes for converting the ground truth class label stored in $\mathcal{GT}$ into one-hot format, $\sum_{i, j}$ denotes that the summation is carried out over all the pixels of the $\mathcal{GT}$, and $\mathcal{L}_{ce}$ is the cross-entropy loss. We empirically set $\alpha = 0.4$ (check appendix for more details).

%% file: parts/04_experiments.tex
\section{Experiments}
\label{sec:experiments}

We compare our approach with Swin Transformer~\cite{swin-T}, and Mix-Transformer~\cite{segformer} with extensive experiments to demonstrate the effectiveness of the SeMask framework. We also ablate the SeMask structure and confirm that providing a semantic-prior to mask out the features improves semantic segmentation performance. The experiments are performed on two widely used datasets: ADE20K \cite{ade20k} and Cityscapes \cite{cityscapes}. We include more experimental results in the appendix proving that our method is dataset agnostic.

\subsection{Datasets and metrics}

\noindent 
\textbf{ADE20K.} \cite{ade20k} ADE20K is a scene parsing dataset covering 150 fine-grained semantic concepts and it is one of the most challenging semantic segmentation datasets. The training set contains 20,210 images with 150 semantic classes. The validation and test set contain 2,000 and 3,352 images respectively.

\noindent
\textbf{Cityscapes.} \cite{cityscapes} Cityscapes is an urban street driving dataset for semantic segmentation consisting of 5,000 images from 50 cities with 19 semantic classes. There are 2,975 images in the training set, 500 images in the validation set and 1,525 images in the test set.

\noindent 
\textbf{Metrics.} We report mean Intersection-over-Union ($mIoU$) over all classes.

\subsection{Implementation details}

\noindent \textbf{Transformer models.} For the encoder, we build upon the Swin Transformer \cite{swin-T} and consider the \textit{Tiny}, Small, \textit{Base} and \textit{Large} variants as described in \cref{tab:swin_desc}. The variation in \textit{number of parameters} among the baselines is due to the \textit{number of transformer blocks} ($N_{T_B}$) (\cref{fig:swin-block}) and the \textit{embedding dimension} ($C$) for each stage of the model. The \textit{number of heads} ($N_{T_H}$) of a shifted window based multi-headed self-attention (SW-MSA) or Swin Transformer block varies from stage to stage. The hidden size of the MLP following SW-MSA is four times the embedding dimension at the corresponding stage. We also experiment with the MiT-B4 backbone variant of the Mix-Transformer~\cite{segformer} on the ADE20K~\cite{ade20k} dataset.

In the following sections, we use an abbreviation to describe the model variant. For example, Swin-T denotes the \textit{Tiny} variant. The backbones pretrained on ImageNet-22k \cite{imagenet} and with $384\!\times\!384$ resolution are denoted with a \dag: Swin-B$^\dag$. All the other models are pretrained on ImageNet-1k and with $224\!\times\!224$ resolution.

\begin{table}
  \centering
  \fontsize{10}{12}\selectfont
  \resizebox{1.\linewidth}{!}{
  \input{tables/swin_desc.tex}}
  \vspaceundertab
  \caption{\textbf{Details of Swin Transformer variants.} The \textit{Tiny} and \textit{Small} variants are trained on ImageNet-1k and with $224\!\times\!224$ resolution. $\dag$ stands for ImageNet-22k pre-training on $384\!\times\!384$ resolution images.}
    \label{tab:swin_desc}
\end{table}

\noindent \textbf{Network Initialization.} Our SeMask models are 
initialized with publicly available models. The \textit{Tiny} and \textit{Small} variants are pre-trained on ImageNet-1k with an image resolution of $224\!\times\!224$. The \textit{Base} and \textit{Large} variants are pretrained on ImageNet-22k with a resolution of $384\!\times\!384$. We keep the window size $(M)$ fixed as in the pretrained models and fine-tune the models for the semantic segmentation task at higher resolution depending on the dataset. Following \cite{swin-T}, we include relative position bias while calculating the attention scores. The decoders, described in \cref{sec:decoder} are initialized with random weights from a normal distribution \cite{init}.

\noindent \textbf{Data augmentation.} During training, we perform mean subtraction, scaling the image to a ratio randomly sampled from $\left ( 0.5, 0.75, 1.0, 1.25, 1.5, 1.75 \right )$, random left-right flipping, and color jittering. We randomly crop large images and pad small images to a fixed size of $512\!\times\!512$ for ADE20K and $768\!\times\!768$ for Cityscapes. On ADE20K, we train our largest model Semask-L$^\dag$ FPN with a $640\!\times\!640$ resolution, matching the resolution used by the Swin-Transformer \cite{swin-T}.

\noindent \textbf{Training Settings.} To fine-tune the pre-trained models on the semantic segmentation task, we employ the AdamW \cite{adamW} optimizer with a base learning rate $\gamma_{0}$. Following the seminal work of DeepLab \cite{deeplabv1} we adopt the \textit{poly} learning rate decay $\gamma = \gamma_0\,(1-\frac{N_{iter}}{N_{total}})^{0.9}$ where $N_{iter}$ and $N_{total}$ represent the current iteration number and the total iteration number. We use a linear warmup strategy for 1,500 iterations.

For ADE20K, we set the base learning rate $\gamma_0$ to $10^{-4}$, weight decay to ${10^{-4}}$ and train for $80K$ iterations with a batch size of $16$.

For Cityscapes, we set $\gamma_0$ to $10^{-3}$, a weight decay of $5 \times {10^{-2}}$ and train for $80K$ iterations with a batch size of $8$.

\noindent \textbf{Inference.} To handle varying image sizes during inference, we keep the aspect ratio intact and resize the image to a resolution with the smaller edge resized to the training resolution and consequently rescaled to the original dimensions before calculating the metric score. For multi-scale inference, following standard practice \cite{deeplabv3+} we use rescaled versions of the image with scaling factors of $\left ( 0.5, 0.75, 1.0, 1.25, 1.5, 1.75 \right )$.

\subsection{Ablation Studies}

In this section, we ablate different variants of our SeMask framework. We investigate the model size, semantic attention, number of \textit{SeMask} blocks ($N_S$), effect of the learnable scalar constant ($\lambda$) inside the \textit{SeMask} block and the pretraining dataset as well as image resolution. Unless stated otherwise, we use the Semantic-FPN \cite{sem-fpn} as our decoder for the main prediction and report results using single-scale (s.s.) inference on the ADE20K \cite{ade20k} val dataset.

\noindent
\textbf{Transformer size.} We study the impact of transformers size on performance in \cref{tab:ablation_model_size} by experimenting with the four different Swin variants: \textit{Tiny}, \textit{Small}, \textit{Base} and \textit{Large} with $N_S = [1, 1, 1, 1]$ for all the experiments. 
Our method gives improvement consistently over all the baseline variants with the improvement on the Cityscapes dataset being more impressive due to the fewer number of classes in the segmentation dataset creating a stronger prior.

We evaluate and record the mIoU scores for the baseline Swin models by training our networks using their publicly released code based on the MMSegmentation Library \cite{mmseg}.

\begin{table*}
  \centering
  \fontsize{10}{12}\selectfont
  \resizebox{1.\linewidth}{!}{
  \input{tables/ablation_model_size.tex}}
  \vspaceundertab
  \caption{\textbf{Ablation on Swin-Transformer varaints.} We provide a comparison of using SeMask Swin with Semantic-FPN \cite{sem-fpn} decoder on all 4 varaints on the \textbf{ADE20K-Val} and \textbf{Cityscapes-Val} dataset. We evaluate the models using both, the \textit{single scale (s.s)} and \textit{multi-scale (m.s.)} mIoU~($\uparrow$). All models are trained for $80k$ iterations. The FLOPs are calculated for the given crop sizes using the script provided by the MMSegmentation \cite{mmseg} library.
      }
    \label{tab:ablation_model_size}
\end{table*}

\noindent
\textbf{Semantic Attention.} We study the impact of the semantic attention operation calculated inside the \textit{SeMask Block} on performance in \cref{tab:ablation_attention} by replacing the SeMask Block with a simple single-head self-attention block on the Swin-Tiny variant. It is evident that simple attention does not help improve the results proving the validity and effectiveness of our SeMask Block.

\begin{table}
  \centering
  \fontsize{10}{12}\selectfont
  \resizebox{1.\linewidth}{!}{
  \input{tables/ablation_attention.tex}}
  \vspaceundertab
  \caption{\textbf{Ablation on Semantic Attention.} 
        We prove the effectiveness of the \textit{SeMask Block} by replacing it with a simple Single-Head Self Attention block which harms the performance on the \textit{Tiny} variant.}
    \label{tab:ablation_attention}
\end{table}

\noindent
\textbf{Learnable Constant ($\lambda$).} We study the impact of $\lambda$ on performance in \cref{tab:ablation_lambda}, by removing it for the \textit{Tiny} and \textit{Small} variants. We observe that the inclusion of $\lambda$ is potent to the success of the SeMask block as it acts as a tuning factor for the modified features, keeping the noise from weights' initialization in check. We also observe that $\lambda \in [0.05, 0.3]$ during inference for different stages in the encoder.

\begin{table}
  \centering
  \fontsize{10}{12}\selectfont
  \resizebox{1.\linewidth}{!}{
  \input{tables/ablation_lambda.tex}}
  \vspaceundertab
  \caption{\textbf{Ablation on $\lambda$.} 
        We support the critical claim of the learnable scalar constant: $\lambda$ inside the SeMask Block by removing and recording the mIoU~($\uparrow$).}
    \label{tab:ablation_lambda}
\end{table}

\noindent
\textbf{Number of SeMask Blocks ($N_S$).} In \cref{tab:ablation_ns} we study the impact of number of SeMask attention blocks on performance by changing the values of $N_S$ inside each semantic layer on the Swin-Tiny variant. We observe that $N_S = [1,1,1,1]$ is the best setting. 
Interestingly, when stacking multiple blocks in a layer, we observe that inputting the $S_Q$ from the previous SeMask block into the later one gives better performance than obtaining $S_Q$ from the features. This shows that extracting semantic features using a single semantic attention operation is the optimum setting.

\begin{table}
  \centering
  \fontsize{10}{12}\selectfont
  \resizebox{1.\linewidth}{!}{
  \input{tables/ablation_ns.tex}}
  \vspaceundertab
  \caption{\textbf{Ablation on $N_S$.} 
        We experiment with different combinations of $N_S$ on the SeMask-Tiny variant and report mIoU~($\uparrow$). $N_S = [1,1,1,1]$ is the best setting.}
    \label{tab:ablation_ns}
\end{table}

\textbf{Pretraining Dataset.} We study the impact of the pretraining dataset (ImageNet-1k {v/s} ImageNet-22k) on performance in \cref{tab:ablation_pretrain} by training and evaluating the \textit{Base} variant pretrained on various settings. Our framework is agnostic to the pretraining setting showing improvement for all combinations mainly used for the ImageNet pretraining: \textit{(i)} ImageNet-1k and $224\!\times\!224$ image resolution; \textit{(ii)} ImageNet-22k and $224\!\times\!224$ image resolution; and \textit{(iii)} ImageNet-22k and $384\!\times\!384$ image resolution.

\begin{table}
  \centering
  \fontsize{10}{12}\selectfont
  \resizebox{1.\linewidth}{!}{
  \input{tables/ablation_pretrain.tex}}
  \vspaceundertab
  \caption{\textbf{Ablation on Pretraining dataset.} 
       We compare the improvement when using the SeMask-Base variant with different pretraining settings: \textit{ImageNet-1k} $v/s$ \textit{ImageNet-22k} and $224\!\times\!224$ $v/s$ $384\!\times\!384$ and show that it is agnostic to the pretraining setting.}
    \label{tab:ablation_pretrain}
\end{table}

\subsection{Main Results}

%In this section, we compare the performance of our SeMask framework to the state-of-the-art methods on the ADE20K and Cityscapes datasets.

\begin{figure*}[ht!]
\centering
\includegraphics[width=0.90\textwidth]{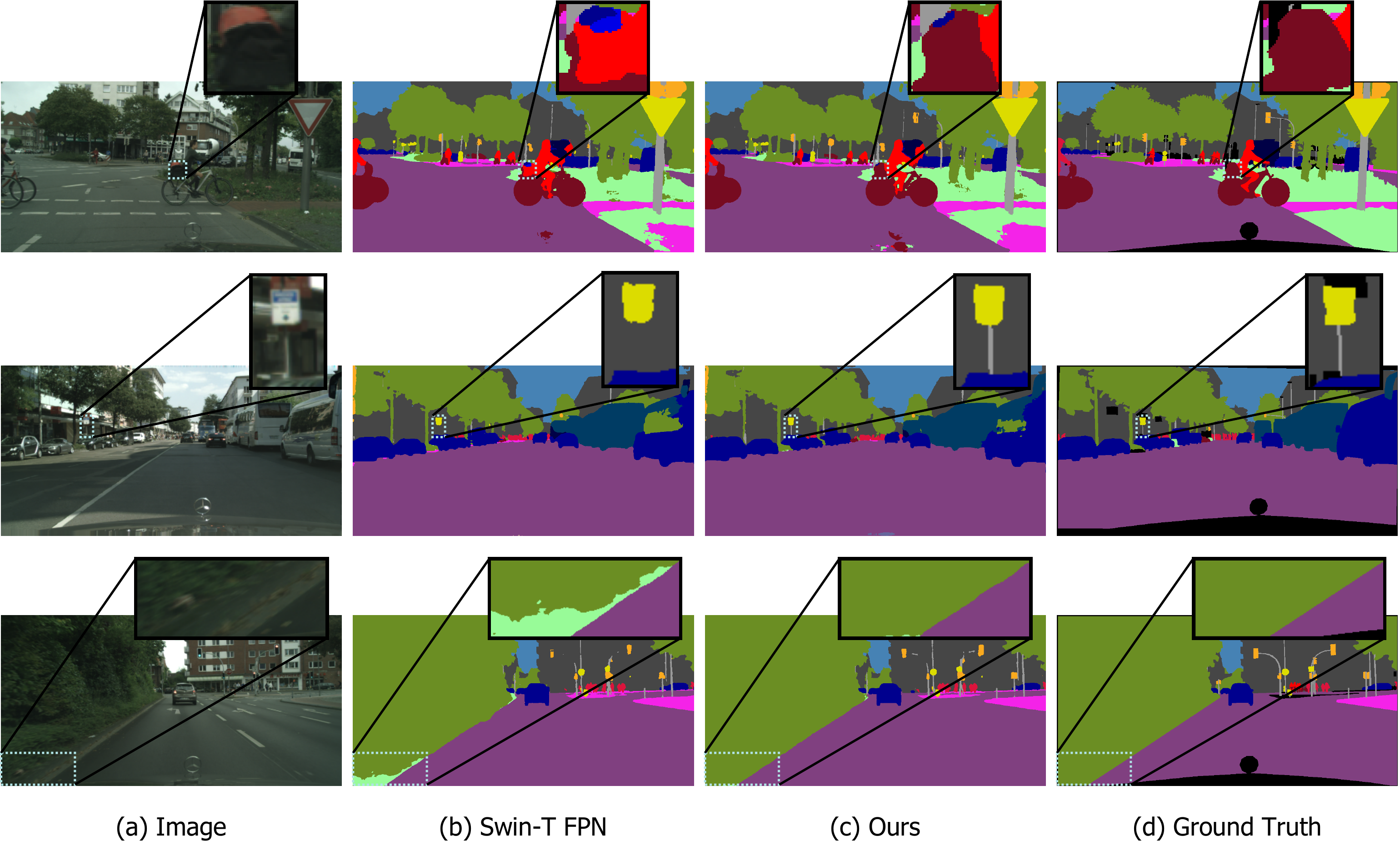}
\caption{
   \textbf{Qualitative results on the Cityscapes validation set.} The dot-bordered boxes at the top show zoomed-in regions from the images for a more detailed look at the improvement using our \textbf{SeMask-T FPN}.}
  \label{fig:qual_cityscapes}
\end{figure*}

\noindent
\textbf{ADE20K.} 
Using SeMask Swin-L$^\dag$ as the encoder and Mask2Former-MSFaPN as our decoder for the main prediction, we achieve a new \textit{state-of-the-art} performance with scores of $57.00\%$ and $58.25\%$ on the single-scale and multi-scale mIoU metric, respectively. Following \cite{swin-T}, our models were trained on $640\!\times\!640$ images. We also achieve competitive results with our SeMask Swin-L$^\dag$ backbone with Semantic-FPN to the Swin-L$^\dag$ based UPerNet model as shown in \cref{tab:sota_ade20k}.

We also integrate our SeMask into the MiT-B4 based SegFormer model~\cite{segformer} as shown in \cref{tab:sota_ade20k} and achieve an improvement of $1.55\%$ on the single scale mIoU and $1.31\%$ improvememt on the multi-scale mIoU metric scores. This supports our claim that SeMask can be plugged into any existing hierarchical vision transformer and show performance improvement.

\begin{table}
  \centering
  \fontsize{10}{12}\selectfont
  %\tablestyle{5pt}{1.2}
  \resizebox{1.\linewidth}{!}{
  \input{tables/sota_cityscapes.tex}}
  \vspace{-2mm}
  \caption{\textbf{SOTA Comparison on Cityscapes-Validation.} 
      We report both single-scale (s.s.) and multi-scale (m.s.) mIOU~($\uparrow$) on \textit{Cityscapes Validation set}.}
    \label{tab:sota_cityscapes}
\end{table}

\begin{table*}
  \centering
  \fontsize{10}{12}\selectfont
  \resizebox{0.95\linewidth}{!}{
  \input{tables/sota_ade20k.tex}}
  \vspaceundertab
  \caption{\textbf{SOTA Comparison on ADE20K-Val.} 
      We report both single-scale (s.s.) and multi-scale (m.s.) mIOU~($\uparrow$) on \textit{ADE20K Val set}. $^\ddagger$ We use the results from the official MMSegmentation~\cite{mmseg} retrained model. $^\|$We develop an MSFaPN network based on the changes done in FaPN~\cite{fapn} to the BasePixelDecoder~\cite{maskformer}. We add similar feature alignment modules based on deformable convolutions~\cite{deform-conv} to the MSDeformAttnPixelDecoder proposed in \cite{mask2former} to obtain the MSFaPN design. $^*$Note that we follow the convention and compare methods based on the Swin-L backbone and we currently do not consider giant models like SwinV2-G that have billions of parameters.}
    \label{tab:sota_ade20k}
\end{table*}

\noindent
\textbf{Cityscapes.} \cref{tab:sota_cityscapes} reports the performance of SeMask on Cityscapes. Semask Swin-L$^\dag$ is competitive with other \textit{state-of-the-art methods} with SeMask Swin-L$^\dag$ Mask2Former achieving $84.98\%$ mIoU. We train our SeMask-L Mask2Former on $512\!\times\!1024$ images following Mask2Former~\cite{mask2former}.

\noindent
\textbf{Qualitative results.}
\cref{fig:qual_cityscapes} shows a qualitative comparison of Swin-T FPN and SeMask-T FPN on the Cityscapes dataset generated using the MMSegmentation library \cite{mmseg}. It is evident that SeMask-T FPN is able to generate better class-wise predictions than the Swin-T FPN. As shown in the second row in \cref{fig:qual_cityscapes}, we are able to segment the pole with out SeMask-T FPN, while Swin-T FPN fails to do so. Similarly in the third row, we are better able to segment the boundary.

%% file: tables/swin_desc.tex
\begin{tabular}{l|ccccc}
    \toprule
    \textbf{Backbone} & \textbf{Window Size} & \textbf{Embedding Dim ($C$)} & \textbf{Blocks ($N_{T_B}$)} & \textbf{Heads ($N_{T_H}$)} & \textbf{\#Params (M)} \\
    \midrule
    Swin-T\phantom{$^\dag$} & 7 & [96, 192, 384, 768] & [2, 2, 6, 2] & [3, 6, 12, 24] & 28 \\
    Swin-S\phantom{$^\dag$} & 7 & [96, 192, 384, 768] & [2, 2, 18, 2] & [3, 6, 12, 24] & 50 \\
    % Swin-B & 7 & [128, 256, 512, 1024] & [2, 2, 18, 2] & [4, 8, 16, 32] & 88 \\
    Swin-B$^\dag$ & 12 & [128, 256, 512, 1024] & [2, 2, 18, 2] & [4, 8, 16, 32] & 88 \\
    Swin-L$^\dag$ & 12 & [192, 384, 768, 1536] & [2, 2, 18, 2] & [6, 12, 24, 48] & 197 \\
    \bottomrule
\end{tabular}

%% file: tables/ablation_model_size.tex
\begin{tabular}{ll|ccccc|ccccc}
   \toprule
   \textbf{Method} & \textbf{Backbone} & \multicolumn{5}{c|}{\textbf{ADE20K}} & \multicolumn{5}{c}{\textbf{Cityscapes}} \\
    (encoder + decoder) & (pretrained) & crop size & \#param. (M) & FLOPs (G) & s.s. mIoU (\%) & m.s. mIoU (\%) & crop size & \#param. (M) & FLOPs (G) & s.s. mIoU (\%) & m.s. mIoU (\%) \\
    \midrule
    
    Swin-T FPN & Swin-T & $512 \times 512$ & $33$ & $38$ & 41.48 \phantom{(+0.00)} & 42.89 \phantom{(+0.00)} & $768 \times 768$ & $33$ & $81$ & 71.81 \phantom{(+0.00)} & 73.74 \phantom{(+0.00)} \\
    SeMask-T FPN & SeMask Swin-T & $512 \times 512$ & $35$ & $40$ & \textbf{42.06} \textcolor{blue}{(+0.58)} & \textbf{43.36} \textcolor{blue}{(+0.47)} & $768 \times 768$ & $34$ & $84$ & \textbf{74.92} \textcolor{blue}{(+3.11)} & \textbf{76.56} \textcolor{blue}{(+2.82)} \\
    \hline
    \addlinespace[0.2cm]
    
    Swin-S FPN & Swin-S & $512 \times 512$ & $54$ & $61$ & 45.20 \phantom{(+0.00)} & 46.96 \phantom{(+0.00)} & $768 \times 768$ & $54$ & $130$ & 75.19 \phantom{(+0.00)} & 77.68 \phantom{(+0.00)} \\
    SeMask-S FPN & SeMask Swin-S & $512 \times 512$ & $56$ & $63$ & \textbf{45.92} \textcolor{blue}{(+0.72)} & \textbf{47.63} \textcolor{blue}{(+0.67)} & $768 \times 768$ & $56$ & $134$ & \textbf{77.13} \textcolor{blue}{(+1.94)} & \textbf{79.14} \textcolor{blue}{(+1.46)} \\
    \hline
    \addlinespace[0.2cm]
    
    Swin-B FPN & Swin-B$^\dag$ & $512 \times 512$ & $93$ & $103$ & 48.80 \phantom{(+0.00)} & 50.28 \phantom{(+0.00)} & $768 \times 768$ & $93$ & $211$ & 76.54 \phantom{(+0.00)} & 79.05 \phantom{(+0.00)} \\
    SeMask-B FPN & SeMask Swin-B$^\dag$ & $512 \times 512$ & $96$ & $107$ & \textbf{49.35} \textcolor{blue}{(+0.55)} & \textbf{50.98} \textcolor{blue}{(+0.70)} & $768 \times 768$ & $96$ & $217$ & \textbf{77.70} \textcolor{blue}{(+1.16)} & \textbf{79.73} \textcolor{blue}{(+0.68)} \\
    \hline
    \addlinespace[0.2cm]
    
    Swin-L FPN & Swin-L$^\dag$ & $640 \times 640$ & $204$ & $343$ & 50.85 \phantom{(+0.00)} & 52.95 \phantom{(+0.00)} & $768 \times 768$ & $204$ & $444$ & 78.03 \phantom{(+0.00)} & 79.53 \phantom{(+0.00)} \\
    SeMask-L FPN & SeMask Swin-L$^\dag$ & $640 \times 640$ & $212$ & $356$ & \textbf{51.89} \textcolor{blue}{(+1.04)} & \textbf{53.52} \textcolor{blue}{(+0.57)} & $768 \times 768$ & $211$ & $455$ & \textbf{78.53} \textcolor{blue}{(+0.50)} & \textbf{80.39} \textcolor{blue}{(+0.86)} \\
    \bottomrule
\end{tabular}

%% file: tables/ablation_attention.tex
\begin{tabular}{ll|cc|cc}
    \toprule
    \textbf{Method} & \textbf{Backbone} & \textbf{SA Block} & \textbf{SeMask Block} & \textbf{mIoU (\%)} & \textbf{\#Param (M)} \\
    \midrule
    Swin-T FPN & Swin-T & & & 41.48 & 33 \\
    Trans Swin-T FPN & Trans Swin-T & $\checkmark$ & & 41.42 & 36 \\
    SeMask-T FPN & SeMask Swin-T & & $\checkmark$ & \textbf{42.06} & 35 \\
    \bottomrule
\end{tabular}

%% file: tables/ablation_lambda.tex
\begin{tabular}{llc|cc}
    \toprule
    \textbf{Method} & \textbf{Backbone} & \textbf{$\lambda$} & \textbf{mIoU (\%)} & \textbf{\#Param (M)} \\
    \midrule
    SeMask-T FPN & SeMask Swin-T & $\checkmark$ & \textbf{42.06} & 35 \\
    SeMask-T FPN & SeMask Swin-T & $\times$ & 41.11 & 35 \\
    \hline
    \addlinespace[0.1cm]
    SeMask-S FPN & SeMask Swin-S & $\checkmark$ & \textbf{45.92} & 56 \\
    SeMask-S FPN & SeMask Swin-S & $\times$ & 45.00 & 56 \\
    \bottomrule
\end{tabular}

%% file: tables/ablation_ns.tex
\begin{tabular}{llc|cc}
    \toprule
    \textbf{Method} & \textbf{Backbone} & \textbf{N$_S$} & \textbf{mIoU (\%)} & \textbf{\#Param (M)} \\
    \midrule
    SeMask-T FPN & SeMask Swin-T & [1, 1, 1, 1] & \textbf{42.06} & 35 \\
    SeMask-T FPN & SeMask Swin-T & [1, 2, 2, 2] & 40.60 & 37 \\
    SeMask-T FPN & SeMask Swin-T & [2, 2, 2, 2] & 40.09 & 37 \\
    \bottomrule
\end{tabular}

%% file: tables/ablation_pretrain.tex
\begin{tabular}{ll|cc|cc}
    \toprule
    \textbf{Method} & \textbf{Backbone} & \textbf{Pre} & \textbf{Res} & \textbf{mIoU (\%)} & \textbf{\#Param (M)} \\
    \midrule
    Swin-B FPN & Swin-B & 1k & 224 & 45.47 & 93 \\
    SeMask-B FPN & SeMask Swin-B & 1k & 224 & \textbf{45.63} & 96 \\
    \hline
    \addlinespace[0.1cm]
    Swin-B FPN & Swin-B & 22k & 224 & 47.65 & 93 \\
    SeMask-B FPN & SeMask Swin-B & 22k & 224 & \textbf{48.29} & 96 \\
    \hline
    \addlinespace[0.1cm]
    Swin-B FPN & Swin-B & 22k & 384 & 48.80 & 93 \\
    SeMask-B FPN & SeMask Swin-B & 22k & 384 & \textbf{49.06} & 96 \\
    \bottomrule
\end{tabular}

%% file: tables/sota_cityscapes.tex
\begin{tabular}{ll|cc}
    \toprule
    \textbf{Method} & \textbf{Backbone} & \textbf{mIoU (\%)} & \textbf{MS mIoU (\%)}  \\
    \midrule
    \multicolumn{4}{l}{\textit{CNN Backbones}}\\
    \hline
    \addlinespace[0.1cm]
    PSANet~\cite{psanet} & ResNet-101 & 77.94 & 79.05 \\
    DeepLabV3+~\cite{deeplabv3+} & Xception-71 & - & 79.55 \\
    CCNet~\cite{ccnet} & ResNet-101 & 80.50 & 81.30 \\
    \hline
    \addlinespace[0.1cm]
    \multicolumn{4}{l}{\textit{Transformer Backbones}}\\
    \hline
    \addlinespace[0.1cm]
    Seg-L-Mask/16~\cite{segmenter} & ViT-L/16$^\dag$ & 79.10 & 81.30 \\
    Swin-L FPN~\cite{swin-T} & Swin-L$^\dag$ & 78.03 & 79.53 \\
    MaskFormer~\cite{maskformer} & ResNet-101 & 78.50 & 80.30\\
    Mask2Former~\cite{mask2former} & Swin-L$^\dag$ & 83.30 & 84.30\\
    HRNetV2-OCR+PSA~\cite{polarized_hrnet} & HRNetV2-W48$^\dag$ & - & \textbf{86.95}\\
    \hline
    \addlinespace[0.1cm]
    SeMask-L FPN \textit{(Ours)} & SeMask Swin-L$^\dag$ & 78.53 & 80.39 \\
    SeMask-L Mask2Former \textit{(Ours)} & SeMask Swin-L$^\dag$ & \textbf{83.97} & \textbf{84.98} \\
    \bottomrule
\end{tabular}

%% file: tables/sota_ade20k.tex
\begin{tabular}{llc|ccc}
    \toprule
    \textbf{Method} & \textbf{Backbone} & \textbf{Crop Size} & \textbf{mIoU (\%)} & \textbf{MS mIoU (\%)} \\
    % & \textbf{\#Param (M)} \\
    \midrule
    \multicolumn{4}{l}{\textit{CNN Backbones}}\\
    \hline
    \addlinespace[0.1cm]
    FCN~\cite{fcn} & ResNet-101 & $512 \times 512$ & 39.91 & 41.40 \\
    % & 69 \\
    PSPNet~\cite{pspnet} & ResNet-101 & $512 \times 512$ & 44.39 & 45.35 \\
    % & 68 \\
    DLab.v3+~\cite{deeplabv3+} & ResNet-101 & $512 \times 512$ & 45.47 & 46.35 \\
    % & 63 \\
    \hline
    \addlinespace[0.1cm]
    \multicolumn{4}{l}{\textit{Transformer Backbones}}\\
    \hline
    \addlinespace[0.1cm]
    SegFormer~\cite{segformer}$^\ddagger$ & MiT-B4 & $512 \times 512$ & 48.46 & 49.76 \\
    SETR-L MLA~\cite{setr} & ViT-L/16$^\dag$ & $512 \times 512$ & --- & 50.28 \\
    % & 308 \\
    % PVT-L FPN~\cite{pvt} & PVT-L$^\dag$ & $512 \times 512$ & - & 42.10 \\
    % & 308 \\
    Swin-L FPN~\cite{swin-T} & Swin-L$^\dag$ & $640 \times 640$ & 50.85 & 52.95 \\
    % & 204 \\
    Seg-L-Mask/16~\cite{segmenter} & ViT-L/16$^\dag$ & $640 \times 640$ & 51.80 & 53.56 \\
    % & 334 \\
    Swin-L UPerNet~\cite{swin-T} & Swin-L$^\dag$ & $640 \times 640$ & --- & 53.50 \\
    SwinV2-L UPerNet~\cite{swinv2}$^*$ & SwinV2-L$^\dag$ & $640 \times 640$ & --- & 55.90 \\
    % & 234 \\
    Swin-L MaskFormer~\cite{maskformer} & Swin-L$^\dag$ & $640 \times 640$ & 54.10 & 55.60 \\
    % & 212 \\
    Swin-L Mask2Former~\cite{mask2former} & Swin-L$^\dag$ & $640 \times 640$ & 56.10 & 57.30\\
    Swin-L MSFaPN-Mask2Former~\cite{mask2former}$^\|$ & Swin-L$^\dag$ & $640 \times 640$ & 55.99 & 57.69\\
    Swin-L FaPN-Mask2Former~\cite{mask2former} & Swin-L$^\dag$ & $640 \times 640$ & 56.40 & 57.70\\
    \hline
    \addlinespace[0.1cm]
    SeMask SegFormer \textit{(Ours)} & SeMask MiT-B4 & $512 \times 512$ & 50.01 & 51.07 \\
    SeMask-L FPN \textit{(Ours)} & SeMask Swin-L$^\dag$ & $640 \times 640$ & 51.89 & 53.52 \\
    % & 211 \\
    SeMask-L MaskFormer \textit{(Ours)} & SeMask Swin-L$^\dag$ & $640 \times 640$ & 54.75 & 56.15 \\
    % & 219 \\
    SeMask-L Mask2Former \textit{(Ours)} & SeMask Swin-L$^\dag$ & $640 \times 640$ & 56.41 & 57.52\\
    SeMask-L FaPN-Mask2Former$^\|$ \textit{(Ours)} & SeMask Swin-L$^\dag$ & $640 \times 640$ & 56.88 & \textbf{58.25}\\
    SeMask-L MSFaPN-Mask2Former \textit{(Ours)} & SeMask Swin-L$^\dag$ & $640 \times 640$ & \textbf{57.00} & \textbf{58.25}\\
    \bottomrule
\end{tabular}

%% file: parts/05_conclusion.tex
\section{Conclusion}\vspace{-0.15cm}

This paper argues that directly finetuning off-the-shelf pretrained transformer backbone networks as encoders for semantic segmentation does not consider the semantic context tied up with the images. We claim that adding a semantic prior to guide the encoder's feature modeling enhances the finetuning process for semantic segmentation. Furthermore, to support our claim, we propose the SeMask Block, which can be plugged into any existing hierarchical vision transformer and uses a semantic attention operation to capture the semantic context and augment the semantic representation of the feature maps. We train and evaluate the proposed framework building on the Swin-Transformer \cite{swin-T} and Mix-Transformer~\cite{segformer} backbones based networks and show a considerable improvement in the semantic segmentation performance on the Cityscapes and ADE20K dataset, with improvements above $3\%$ on the Cityscapes dataset. We provide a comprehensive experimental analysis applying SeMask to different backbone variants and achieving considerable performance improvement in every setting. Our method also achieves the new state-of-the-art performance on the ADE20K dataset. As a direction for future research, it will be interesting to observe the effect of adding similar priors for other vision downstream tasks like object detection and instance segmentation.

%% file: supplementary/supplementary.tex
% \cleardoublepage
\appendix
\begin{center}{\bf \Large Appendix}\end{center}\vspace{-2mm}
\renewcommand{\thetable}{\Roman{table}}
\renewcommand{\thefigure}{\Roman{figure}}
\setcounter{table}{0}
\setcounter{figure}{0}

\Crefname{appendix}{Appendix}{Appendixes}

\begin{figure*}[ht!]
\centering
\includegraphics[width=\linewidth]{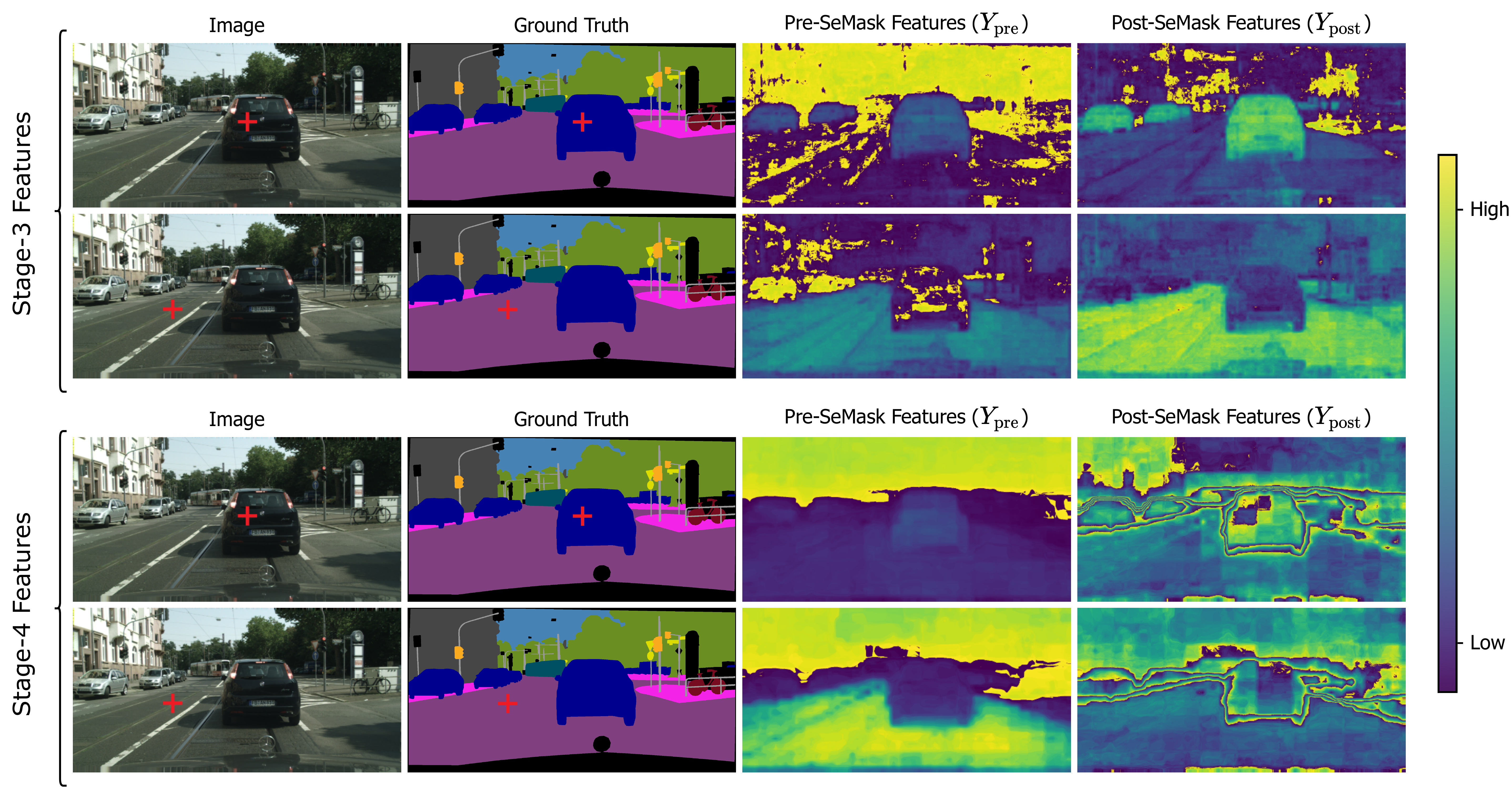}
\caption{
  \textbf{Analysis of features on the Cityscapes val set.} We analyze pixel-wise attention maps for the $Y_\text{pre}$ and $Y_\text{post}$ features from Stage-3 and Stage-4 of our SeMask-T FPN network. The post-SeMask ($Y_\text{post}$) features are richer in clear boundaries and pixel similarity than pre-SeMask ($Y_\text{pre}$) features.}
  \label{fig:analysis}
\end{figure*}

In this appendix, we present an additional ablation study for the value of the weight, $\alpha$ in \cref{sec:alpha}. Then, we share our experimental results with SeMask on the COCO-Stuff 10k dataset in \cref{sec:coco}.  We also provide an analysis on the SeMask's effect on the feature maps in \cref{sec:analysis}. \cref{sec:qual} provides a qualitative comparison of SeMask-L FPN to Swin-L FPN on the COCO-Stuff 10k \cite{coco} and ADE20K \cite{ade20k} datasets. 
% Lastly, we discuss a few limitations of SeMask.

\input{supplementary/weight_ablation}

\input{supplementary/coco_stuff}
\input{supplementary/analysis}
\input{supplementary/qualitative_results}

%% file: supplementary/weight_ablation.tex
\section{Tuning the hyperparameter $\alpha$}
\label{sec:alpha}

We weigh the loss ($\mathcal{L}_{2}$) calculated on the semantic-prior prediction with a hyperparameter $\alpha$ as formulated in \cref{eq:app_loss}. Using weighted supervision for the semantic-prior maps is critical so that the model treats the semantic context as an additional signal for feature modeling and not as the main prediction.

\begin{equation}\label{eq:app_loss}
   \mathcal{L_T} = \mathcal{L}_{1} + \alpha\mathcal{L}_{2}
\end{equation}

We study the impact of $\alpha$ on performance in \cref{tab:ablation_alpha} by changing the values of $\alpha$ on the Swin-Tiny variant. $\alpha = 0.4$ is the optimum setting for modeling the network's image feature level and semantic level context.

\begin{table}[ht]
  \centering
  \resizebox{1.\linewidth}{!}{
  \input{supplementary/tables/ablation_alpha.tex}}
  \vspaceundertab
  \caption{\textbf{Ablation on $\alpha$.} 
        We experiment with different values of $\alpha$ on the SeMask-Tiny variant and report single-scale mIoU~($\uparrow$). $\alpha = 0.4$ is the best setting.}
    \label{tab:ablation_alpha}
\end{table}

%% file: supplementary/tables/ablation_alpha.tex
\begin{tabular}{ll|c|cc}
    \toprule
    \textbf{Method} & \textbf{Backbone} & \textbf{$\alpha$} & \textbf{mIoU (\%)} & \textbf{\#Param (M)} \\
    \midrule
    SeMask-T FPN & SeMask Swin-T & 0.4 & \textbf{42.06} & 35 \\
    SeMask-T FPN & SeMask Swin-T & 0.7 & 41.87 & 35 \\
    SeMask-T FPN & SeMask Swin-T & 1 & 41.67 & 35 \\
    \bottomrule
\end{tabular}

%% file: supplementary/coco_stuff.tex
\section{Experiments on COCO-Stuff 10k}
\label{sec:coco}

COCO-Stuff 10k comprises of a total of 10k images with dense pixel-level annotations, selected from the COCO \cite{coco} dataset. The training set contains 9k images with 171 semantic classes and the test set contains 1k images.

We set the base learning rate $\gamma_0$ to $10^{-4}$, weight decay to ${10^{-4}}$ and train for 80K iterations with a batch size of 16.

We provide our experimental results in \cref{tab:coco_stuff}. Our SeMask framework shows impressive improvement on the COCO-Stuff 10k dataset proving its dataset-agnostic ability.

\begin{table}[ht!]
  \centering
  \small
  \resizebox{1.\linewidth}{!}{
  \input{supplementary/tables/experiments_coco_stuff.tex}}
  \vspaceundertab
  \caption{\textbf{Experiments with COCO-Stuff 10k.} We provide a comparison of using SeMask Swin with Semantic-FPN \cite{sem-fpn} decoder on the COCO Stuff-10k test set. We evaluate the models using both, the \textit{single scale (s.s)} and \textit{multi-scale (m.s.)} mIoU~($\uparrow$). 
      }
    \label{tab:coco_stuff}
\end{table}

%% file: supplementary/tables/experiments_coco_stuff.tex
\begin{tabular}{ll|cc|cc}
   \toprule
   \textbf{Method} & \textbf{Backbone} & \textbf{Crop Size} & \textbf{\#Param. (M)} & \textbf{s.s. mIoU (\%)} & \textbf{m.s. mIoU (\%)} \\
    \midrule
    Swin-T FPN & Swin-T & $512\!\times\!512$ & $33$ & 37.14 \phantom{(+0.00)} & 38.37 \phantom{(+0.00)} \\
    SeMask-T FPN & SeMask Swin-T & $512\!\times\!512$ & $35$ & \textbf{37.53} \textcolor{blue}{(+0.39)} & \textbf{38.88} \textcolor{blue}{(+0.55)} \\
    \hline
    \addlinespace[0.1cm]
    Swin-S FPN & Swin-S & $512\!\times\!512$ & $54$ & 40.53 \phantom{(+0.00)} & 41.91 \phantom{(+0.00)} \\
    SeMask-S FPN & SeMask Swin-S & $512\!\times\!512$ & $56$ & \textbf{40.72} \textcolor{blue}{(+0.19)} & \textbf{42.27} \textcolor{blue}{(+0.36)} \\
    \hline
    \addlinespace[0.1cm]
    Swin-B FPN & Swin-B$^\dag$ & $512\!\times\!512$ & $54$ & 44.18 \phantom{(+0.00)} & 45.79 \phantom{(+0.00)} \\
    SeMask-B FPN & SeMask Swin-B$^\dag$ & $512\!\times\!512$ & $56$ & \textbf{44.68} \textcolor{blue}{(+0.50)} & \textbf{46.30} \textcolor{blue}{(+0.51)} \\
    \hline
    \addlinespace[0.1cm]
    Swin-L FPN & Swin-L$^\dag$ & $640\!\times\!640$ & $204$ & 46.42 \phantom{(+0.00)} & 48.13 \phantom{(+0.00)} \\
    SeMask-L FPN & SeMask Swin-L$^\dag$ & $640\!\times\!640$ & $211$ & \textbf{47.47} \textcolor{blue}{(+1.05)} & \textbf{48.54} \textcolor{blue}{(+0.41)} \\
    \bottomrule
\end{tabular}

%% file: supplementary/analysis.tex
\section{Analysis on SeMask} 
\label{sec:analysis}

In order to confirm our hypothesis that adding semantic context inside the encoder with the help of the semantic attention operation helps in improving the semantic quality of the features, we analyze the pixel-wise attention quality of the intermediate features of our SeMask-T FPN model on the Cityscapes~\cite{cityscapes} val dataset as shown in \cref{fig:analysis}.

\begin{figure}[ht!]
\centering
\includegraphics[width=\linewidth]{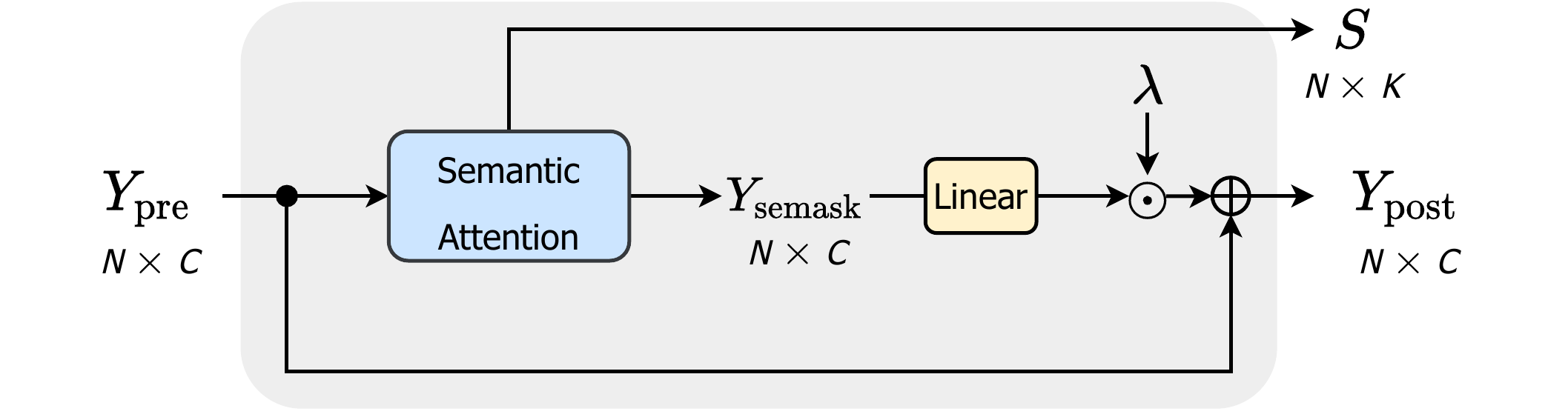}
\caption{
  \textbf{SeMask Block.} The semantic attention outputs the semask features ($Y_\text{semask}$) using the features from the transformer layer ($Y_\text{pre}$). We use a residual connection from $Y_\text{pre}$ to obtain the final output ($Y_\text{post}$). $S$ is the semantic-prior map used to semantically mask the features ($Y_\text{pre}$).}
  \label{fig:supp_semask}
\end{figure}

Specifically, we analyze pixel-wise attention for the pre-SeMask ($Y_\text{pre}$) and post-SeMask ($Y_\text{post}$) features (\cref{fig:supp_semask}) for Stage-3 and Stage-4 which are downsampled by $\times16$ and $\times32$, respectively. We calculate the pixel-wise attention maps corresponding to the target pixel (\textcolor{red}{red} cross sign), and we observe that post-SeMask features have more similar features for the same semantic category region with better boundaries than the pre-SeMask features. It reflects that the semantic prior maps help increase similarity between the pixels belonging to the same semantic category and improve the semantic segmentation performance.

%% file: supplementary/qualitative_results.tex
\section{Qualitative results}
\label{sec:qual}

\begin{figure*}
\centering
\includegraphics[width=.85\linewidth]{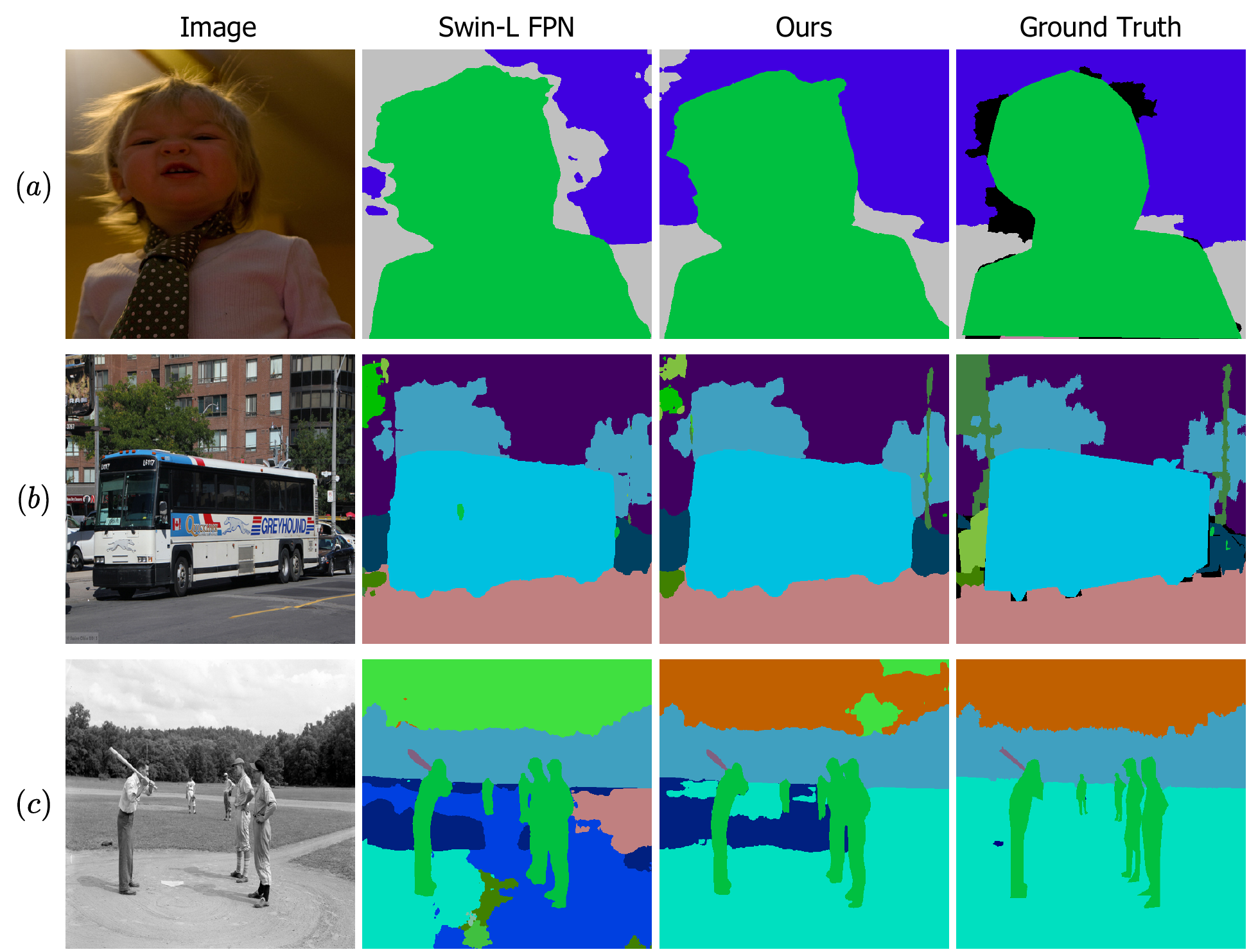}
\caption{
  \textbf{Qualitative results on the COCO-Stuff 10k test set.} Swin-L FPN completely mislabels the sky region and a significant part of the ground in $(c)$, and our \textbf{SeMask-L FPN} shows better accuracy in classifying the regions.}
  \label{fig:qual_coco}
\end{figure*}

We provide qualitative results on the COCO-Stuff 10k test set in \cref{fig:qual_coco} where SeMask-L FPN produces better per-pixel predictions compared to Swin-L FPN. It is evident in $(b)$ as the Swin-L FPN network fails to label the pole correctly and completely mislabels the sky region in $(c)$.

\begin{figure*}
\centering
\includegraphics[width=.85\linewidth]{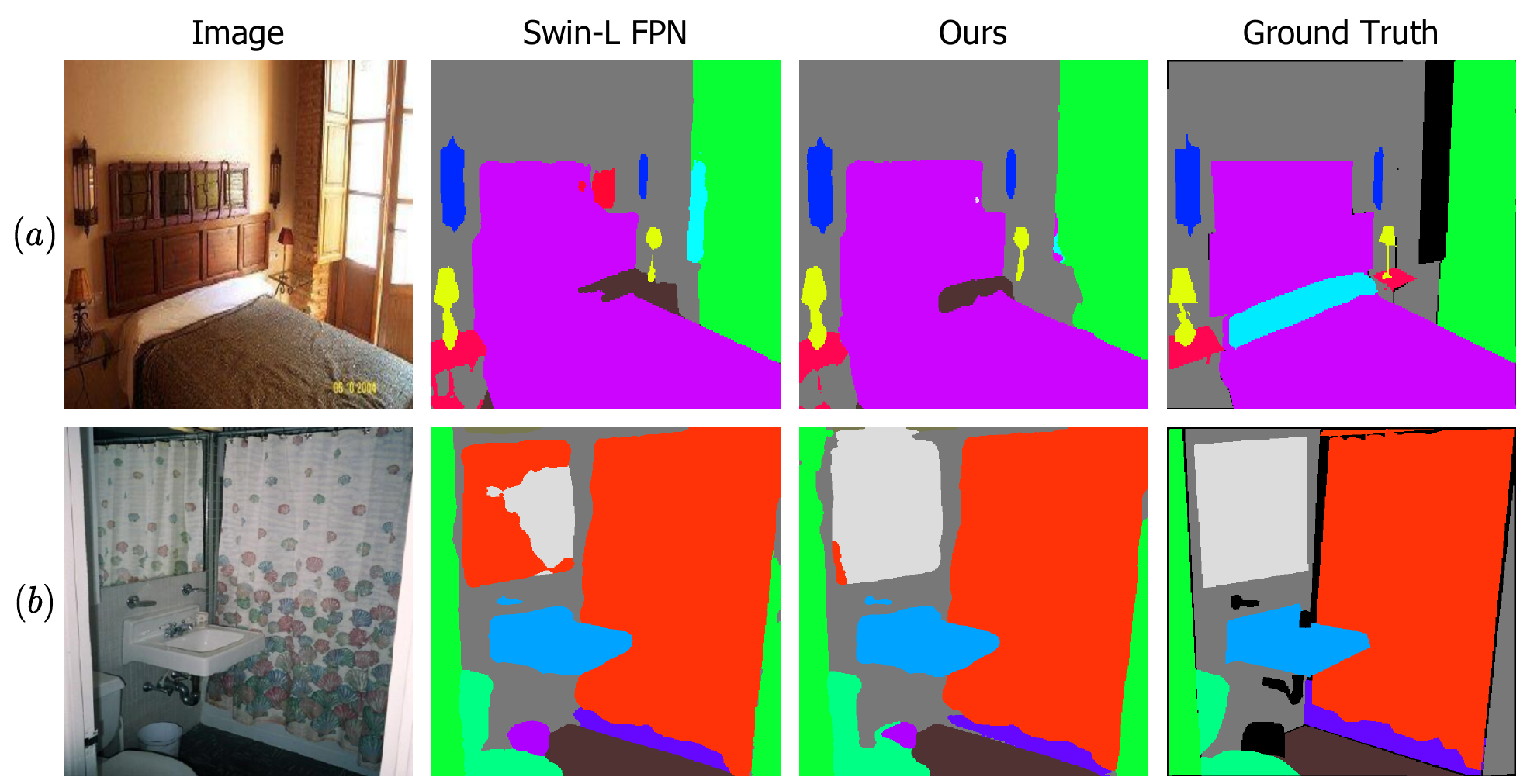}
\caption{
  \textbf{Qualitative results on the ADE20K validation set.} Our \textbf{SeMask-L FPN} can correctly classify the mirror region in $(b)$, whereas Swin-L FPN mislabels a significant part of the mirror as curtain owing to the reflection.}
  \label{fig:qual_ade20k}
\end{figure*}

We show more qualitative results on the ADE20K validation set in \cref{fig:qual_ade20k}. Swin-L FPN mislabels \textit{mirror} as \textit{curtain} in $(b)$ due to the reflection of the curtain. On the other hand, SeMask-L FPN classifies the regions accurately.

